\pdfoutput=1 

\documentclass[preprint,12pt]{elsarticle}

\usepackage{amssymb}
\usepackage[numbers]{natbib}
\usepackage{times}
\usepackage{epsfig}
\usepackage{graphicx}
\usepackage{url}
\usepackage{xcolor}
\usepackage{amsmath}
\usepackage{subfigure}
\usepackage{multirow}
\usepackage{caption}
\usepackage{arydshln}
\usepackage{setspace}

\journal{Nuclear Physics B}

\begin{document}

\begin{frontmatter}



\title{Multi-Scale Local-Temporal Similarity Fusion for Continuous Sign Language Recognition}


\author[inst1]{Pan Xie}

\affiliation[inst1]{organization={Beihang University},
            city={Beijing},
            country={China}}

\author[inst2]{Zhi Cui}
\author[inst1]{Yao Du}
\author[inst1]{Mengyi Zhao}
\author[inst2]{Jianwei Cui}
\author[inst2]{Bin Wang}
\affiliation[inst2]{organization={Xiaomi AI Lab},
            city={Beijing},
            country={China}}

\author[inst3]{Xiaohui Hu}
\affiliation[inst3]{organization={Institute of Software, Chinese Academy of Sciences},
            city={Beijing},
            country={China}}

\begin{abstract}
Continuous sign language recognition (cSLR) is a public significant task that transcribes a sign language video into an ordered gloss sequence. It is important to capture the fine-grained gloss-level details, since there is no explicit alignment between sign video frames and the corresponding glosses. Among the past works, one promising way is to adopt a one-dimensional convolutional network (1D-CNN) to temporally fuse the sequential frames. However, CNNs are agnostic to similarity or dissimilarity, and thus are unable to capture local consistent semantics within temporally neighboring frames. To address the issue, we propose to adaptively fuse local features via temporal similarity for this task. Specifically, we devise a \textbf{M}ulti-scale \textbf{L}ocal-\textbf{T}emporal \textbf{S}imilarity \textbf{F}usion Network (mLTSF-Net) as follows: \textbf{1)} In terms of a specific video frame, we firstly select its similar neighbours with multi-scale receptive regions to accommodate different lengths of glosses. \textbf{2)} To ensure temporal consistency, we then use position-aware convolution to temporally convolve each scale of selected frames. \textbf{3)} To obtain a local-temporally enhanced frame-wise representation, we finally fuse the results of different scales using a content-dependent aggregator. We train our model in an end-to-end fashion, and the experimental results on RWTH-PHOENIX-Weather 2014 datasets (RWTH) demonstrate that our model achieves competitive performance compared with several state-of-the-art models.
\end{abstract}




\end{frontmatter}


\section{Introduction}\label{intro}
Sign language (SL), consisted of various hand gestures, motions, facial expressions, transitions and etc, is an indispensable daily communication tool between people with hearing impairment. A gloss is the minimal semantic unit of sign language~\cite{Ong2005AutomaticSL}, and people would have to accomplish one single or a set of gestures in order to sign a gloss.  

The video-based continuous Sign Language Recognition (cSLR) is the task that transcribes a sequence of signing frames into an ordered gloss sentence. However, this task is challenging: \textbf{1)} On the one hand, many gestures or actions could be shared across different glosses due to the richness of the vocabulary in a sign language. \textbf{2)} On the other hand, even a same gloss may also be comprised of different lengths of corresponding video frames, as different people would act in different speed. These aspects address difficulties in explicitly aligning the video frames with the corresponding signing glosses. Therefore, it would be challenging to recognize glosses from such a dynamic sequential video, where there is no explicit semantic boundaries between glosses.

\begin{figure}[t]
\centering
\includegraphics[width=0.8\textwidth]{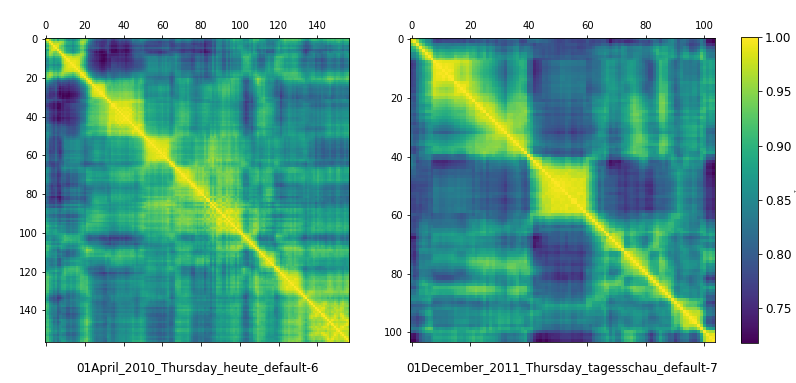}
\caption{Case study of two examples: the figures above respectively visualize the pairwise cosine similarities $cos_{ij}$ between the spatial features of frame$_{i}$ and frame$_{j}$, where the diagonal elements always equal 1. Note that  spatial features are extracted from a CNN backbone model. }
\label{temporal similarity visual} 
\end{figure}

To address the \textit{aforementioned} challenge, many recent approaches have achieved excellent performance by adopting a mainstream framework~\cite{Yang2019SFNetSF,Cheng2020FullyCN, Zhou2020SpatialTemporalMN}: they firstly utilize a convolution neural network (CNN) to extract a sequence of spatial features from video frames, and then \textit{fuse} the temporal sequence either using a recurrent neural network (RNN) or one-dimensional convolution neural network (1D-CNN). However, these two conventional \textit{fusion} methods still have several problems. In terms of RNNs, previous research~\cite{Cheng2020FullyCN} has indicated that they tend to focus more on the sequential order of sign language than the fine-grained gloss-level details, and thus RNNs are not suitable for this task where the local details play a vital role.

On the contrary, it is more promising to use the fully convolutional neural network (FCN)~\cite{Cheng2020FullyCN}, which uses 1D-CNNs to learn such gloss-level granularities. However 1D-CNNs are content-agnostic~\cite{Su2019pixel}, i.e., the same trained CNN filters are applied to all the videos frames without distinction between locally similar or dissimilar ones. But note that the sign language is semantically coherent, suggesting that the temporally close video frames share similarities~\cite{Li2020tspnet}. To further demonstrate such phenomenon, we plot similarity heatmaps of two examples shown in Figure~\ref{temporal similarity visual}. We can clearly see that the temporal neighbours share much higher cosine similarities than those far-away ones, where there are more yellow pixels along the diagonal of the heatmaps. Besides, another research~\cite{Islam2020HowMP} has implied that a learned CNN filter could also be insensitive to where it is targeting. As a result, 1D-CNNs would tend to skip temporal consistency of the convolved frames. To sum up, 1D-CNNs alone could not model this task very well if without extra regulations. 

\begin{figure}
\centering
\includegraphics[width=0.8\textwidth]{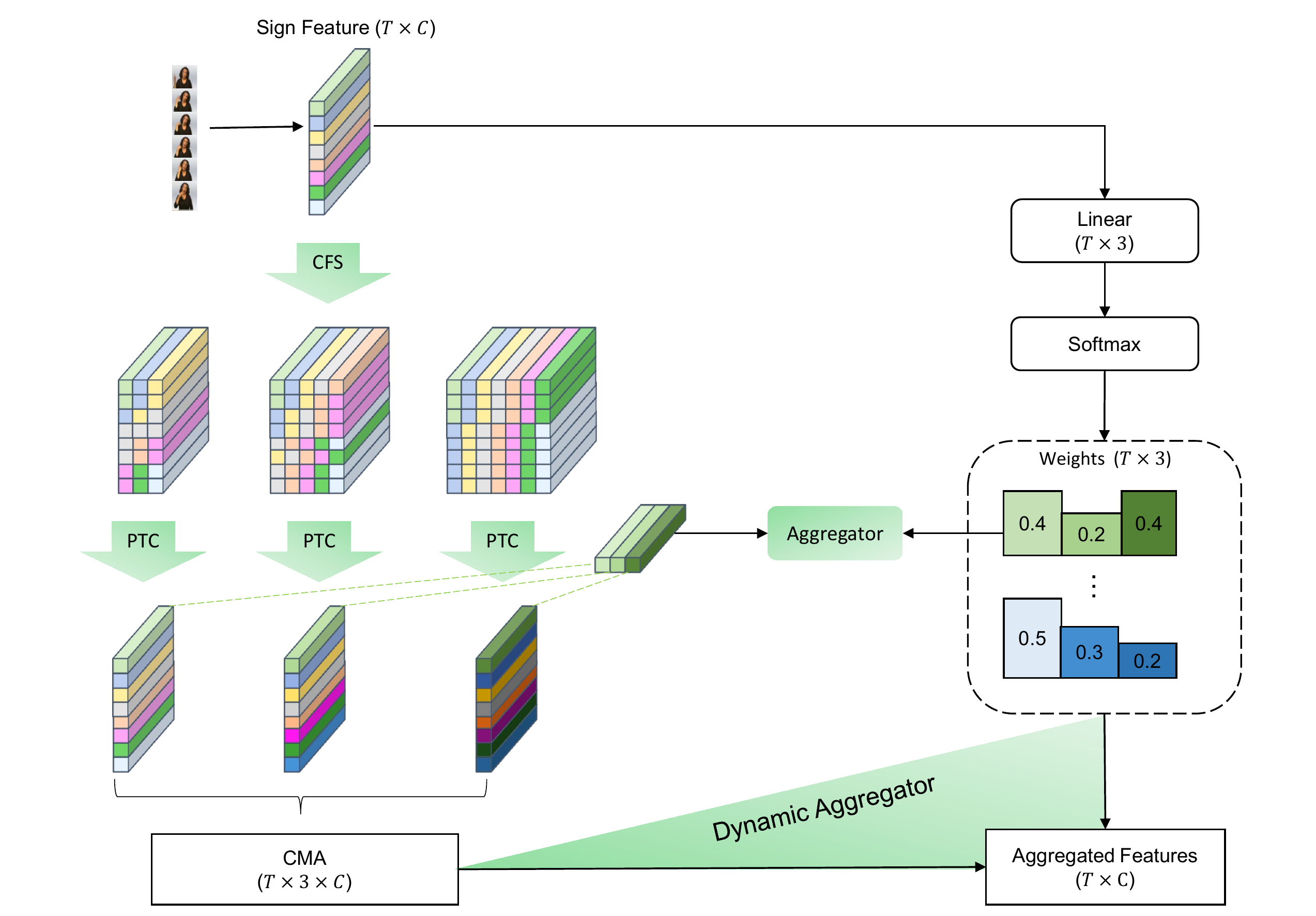}
\caption{An overview of our model, where we mainly zoom on in our proposed mLTSF module. In this figure, "CFS" stands for content-aware feature selector, "PTC" stands for position-aware temporal convolver, and "CMA" stands for content-dependent multi-scale aggregator.}
\label{mtsf} 
\end{figure}

To tackle the problem, we propose to adaptively augment the sequential \textit{fusion} with locally similar frames for continuous Sign Language Recognition. Specifically, we introduce a (\textbf{m}ulti-scale \textbf{L}ocal-\textbf{Temporal} \textbf{S}imilarity \textbf{F}usion) \textbf{Net}work (mLTSF-Net). Compared with the fully convolutional neural network (FCN)~\cite{Cheng2020FullyCN}, our model additionally has an mLTSF module as shown in Figure~\ref{mtsf}, which brings the following improvements: \textbf{1)} Rather than a fixed receptive region, we propose a content-aware feature selector (CFS) which firstly selects similar neighboring frames with different scaled selection regions. Facilitated by this, our model is induced to accommodate different lengths of glosses. \textbf{2)} To enhance temporal consistency,  we then introduce a position-aware temporal convolver (PTC), where each scale of selected frames will be appended with relative positional embeddings and convolved in an temporal order. \textbf{3)} To augment local-temporal frame-wise representations, we finally \textit{fuse} different scales of convolved features with a content-dependent multi-scale aggregator (CMA), where we incorporate a dynamic weight using an attention-like mechanism. We conduct experiments on the benchmark dataset, RWTH-PHOENIX-Weather-2014 (RWTH)~\cite{Forster2014ExtensionsOT}. We also apply extensive ablation studies to validate the effectiveness of our model. The experimental results demonstrate that our proposed model mLTSF-Net achieves the state-of-the-art accuracy (with a word error rate (WER) as 23.0$\%$) compared with many competitive baseline models.

\section{Related Work}\label{related work}
Most researches on continuous sign language recognition mainly focus on two aspects (feature-level and sequence-level): \textbf{1)} either strive to better model the spatial or temporal semantic information of sign language video frames, or \textbf{2)} design a better sequence learning mechanism to learn the mapping relationship between the sign video frames and the corresponding gloss sequences.


\paragraph{\textbf{Feature-Level}} Generally, the feature-level researches for cSLR include two aspects: spatial modeling and temporal modeling. Early works of cSLR mostly use the hand-crafted feature for sign video representation \cite{nguyen2012facial,Starner1998RealTimeAS, Habili2004SegmentationOT,Wang2014SimilarityAM,kong2014towards}. Recently, deep learning based methods achieve the state-of-the-art in cSLR for their strong feature representation capability\cite{Koller2016DeepHH,kumar2017multimodal,Camgz2017SubUNetsEH, Cui2017RecurrentCN,Cui2019ADN,Cheng2020FullyCN,Zhou2020SpatialTemporalMN,gao2021rnn}. A common practice is to use 2D-CNNs to extract the spatial features, and then RNNs or 1D-CNNs are employed to model the temporal variations of sign language. Cui \textit{et al.}~\cite{Cui2017RecurrentCN} utilizes the combination of 2D-CNNs and LSTMs to learn the spatial-temporal representation. Cheng \textit{et al.}~\cite{Cheng2020FullyCN} adopts a fully convolutional network which is composed of 2D-CNNs and 1D-CNNs. The comparison of these two approaches shows that 1D-CNNs are more robust than RNNs for cSLR to model the semantic structure of sign glosses.

\paragraph{\textbf{Sequence-Level}} Sequence learning in cSLR is to learn the mapping relationship between sign video sequences and sign gloss labels. Early on, Hidden markov models (HMMs) are widely used to model the state transitions in cSLR~\cite{gao2004chinese,Koller2016DeepHH,Koller2017ReSignRE,Koller2016DeepSH}. Recently, the attention-based encoder-decoder architecture have shown a great success in machine translation~\cite{Bahdanau2015NeuralMT}. Despite that several works show promising results to apply such architecture for cSLR~\cite{Huang2018VideobasedSL, Guo2018HierarchicalLF}, it is time- and labor-consuming to build a very large-scale source-to-target mapping dataset. Moreover, Connectionist Temporal Classification (CTC)~\cite{Graves2006ConnectionistTC} has proven to be the  state-of-the-art performance in cSLR, where CTC algorithm succeeds in processing unsegmented sequence~\cite{Camgz2017SubUNetsEH, Cui2017RecurrentCN,Cui2019ADN, Cheng2020FullyCN}.


\paragraph{\textbf{Ours}} In summary, our model is built upon the fully convolutional network (FCN) ~\cite{Cheng2020FullyCN}. In particular, FCN use 1D-CNNs to learn temporal information of video features with a window sliding over the time dimension. Despite the successful performance, CNNs are agnostic to similar neighbouring frames~\cite{Su2019pixel}, and also tend to be insensitive to temporal dependency of ordered frames~\cite{Islam2020HowMP}. These drawbacks would hinder the performance of model from being further improved. In this paper, our contributions lie in that we propose a \textbf{m}ulti-scale \textbf{L}ocal-\textbf{T}emporal \textbf{Simialrity} \textbf{Fusion} \textbf{Net}work (mLTSF-Net) to address the above issue, and our model achieves the state-of-the-art performance compared with many competitive baseline models.

\begin{figure}
\centering
\includegraphics[width=\textwidth]{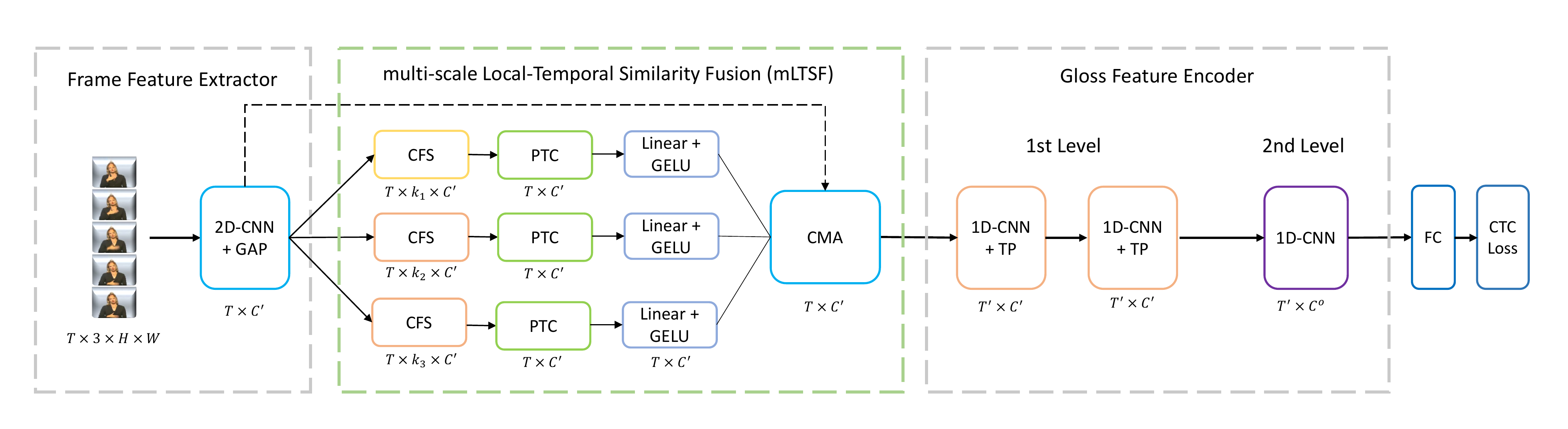}
\caption{An overview of our proposed model. "CFS" means content-aware feature selector. "PTC" means position-aware convolver. "CMA" means content-dependent aggregator. "GELU" means gaussian error linear units activation.} 
\label{arch} 
\end{figure}

\section{Method}\label{method}

\subsection{Preliminaries}
Towards the task of cSLR, our objective is to learn a mapping probability \(p(\mathcal{Y|X})\) from a given series of video frames \(\mathcal{X}=\{x_t\}^T_{t=1}\) to an ordered gloss sequence \(\mathcal{Y}=\{y_i\}^L_{i=1}\), where $T$ and $L$ respectively denote the lengths. 

To model the mapping probability, a promising method is to adopt a fully convolutional network (FCN)~\cite{Cheng2020FullyCN} to learn feature representations. To be specific, the FCN framework is consisted of two main components in a top-down order: \textbf{1)} a frame feature extractor, and \textbf{2)} a gloss feature encoder. Due to the drawback of CNNs as discussed previously, the CNN-stacked gloss feature encoder is insufficient to fully capture gloss-level details.

To address this issue,  we introduce a (\textbf{m}ulti-scale \textbf{L}ocal-\textbf{Temporal} \textbf{S}imilarity \textbf{F}usion) \textbf{Net}work (mLTSF-Net) as shown in Figure~\ref{arch}. Built upon FCN, our proposed model has an additional mLTSF module on top of the gloss feature encoder. In the following subsections, we will firstly revisit the FCN framework and then give a detailed description about our proposal.

\subsection{FCN Framwork}
\paragraph{\textbf{Frame Feature Extractor}} Given a series of video frames $\mathcal{X}=\{x_t\}_{t=1}^{T}$, the frame feature extractor firstly encodes them, and then outputs a series of according spatial features. Specifically, CNNs are commonly adopted as the backbone model to learn the spatial features as:

\begin{equation}
\begin{aligned}
z^{T\times C' \times H' \times W'} = \mathcal{F}_{2d-CNN}(x^{T\times C \times H \times W})
\end{aligned}
\label{eqn:equation1}
\end{equation}

\noindent where \(C, H, W\) respectively denote the number of spatial channels (here $C=3$), the height and weight of the input frame, while \(C', H', W'\) denote the sizes of a new activation feature map $z$ with a lower-resolution. 

To collapse the spatial dimension of $z$, a global average pooling function $\mathcal{F}_{GAP}$ is adopted:

\begin{equation}
\begin{aligned}
s^{T\times C'} = \mathcal{F}_{GAP}(z^{T\times C' \times H' \times W'})
\end{aligned}
\label{eqn:equation2}
\end{equation}

\noindent where $s^{T\times C'}$ denotes the sequence of the frame-wise representations, and \(C'\) is the number output channels. 

\paragraph{\textbf{Gloss Feature Encoder}} Given the extracted spatial features, the gloss feature encoder adopts 1D-CNNs over the time dimension to learn contextual information. Such encoder is consisted of two levels, where each level is stacked with multiple 1D-CNN layers and undertakes different functions: 

\textbf{The 1st Level.} To capture the semantic information of a gloss in terms of the whole sign video, the \textbf{1st level} gloss feature encoder adopts a larger filter size in order to consider more frames at a time. Since it is hard to  learn extremely long dependencies in a temporal sequence~\cite{Cui2019ADN}, temporal max pooling layers are utilized to pool the feature sequence to a moderate length. In other words, the stacked temporal layers can be regarded as a window sliding function over the time dimension. Mathematically, we obtain the gloss feature from the 1st level as:

\begin{equation}
\begin{aligned}
g_{1}^{T' \times C'} = \mathcal{F}_{TP}(\mathcal{F}_{1d-CNN}(s^{T\times C'}))
\end{aligned}
\label{eqn:equation3}
\end{equation}

\noindent where $T'$ is smaller than $T$ due the temporal pooling.

\textbf{The 2nd Level.} To learn the neighboring information in between glosses, the \textbf{2nd level} of gloss feature encoder uses a relatively smaller filter size to emphasize on the local details. The 2nd level does not has a temporal pooling layer, where the output time dimension would not be changed. Likewise, the output gloss feature from 2nd level is represented as:

\begin{equation}
\begin{aligned}
g_{2}^{T'\times C^{o}} = \mathcal{F}_{1d-CNN}(g_{1}^{T'\times C'})
\end{aligned}
\label{eqn:equation4}
\end{equation}

\noindent where $C^{0}$ is the output feature channels of 2nd level. 

\begin{figure*}
\centering
\includegraphics[width=\textwidth]{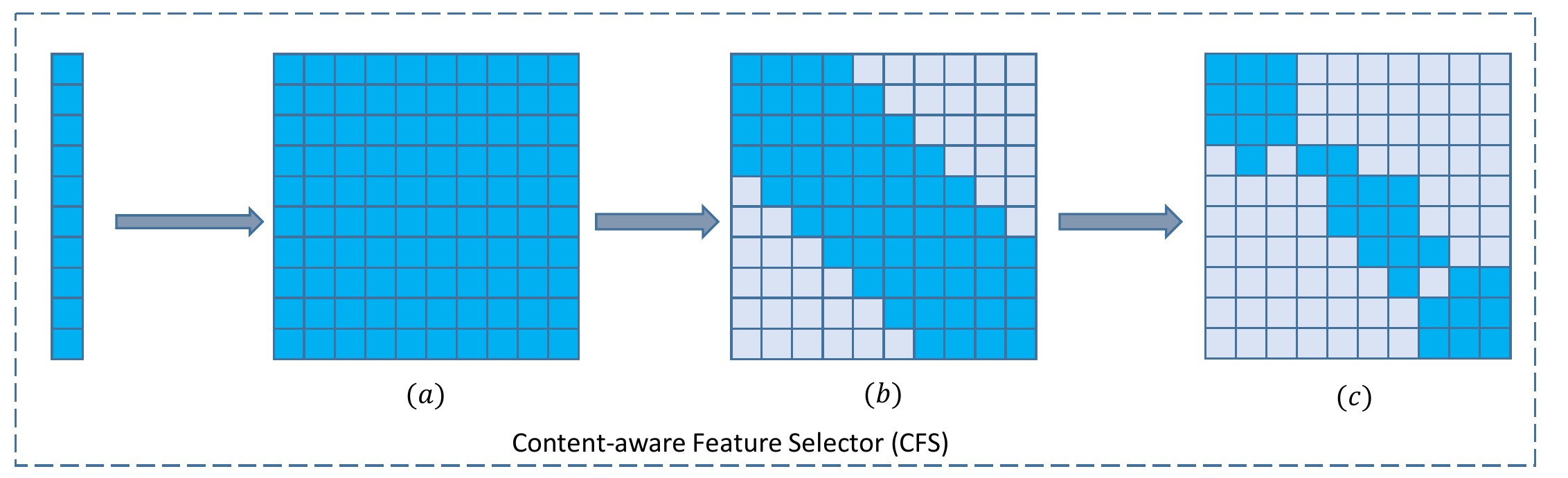}
\caption{An overview of three-step selection by content-aware feature selector.} 
\label{feature_selector} 
\end{figure*}

\subsection{Multi-scale Local-Temporal Similarity Fusion}
It is worth mentioning again that temporal neighboring frames share consistent local semantics~\cite{Li2020tspnet}, which highlights the importance of local-temporal similarity. However, the fully CNN-stacked gloss feature encoder is hard to catch such detailed information, since CNNs are unable to take additional consideration into temporally similar and consistent neighbours. More specific reasons are detailed in Section~\ref{intro}. 

Accordingly, we introduce mLTSF-Net to learn the local-temporal similarity. As mentioned before, our model has an additional mLTSF module compared with FCN. Specifically, our proposed mLTSF module has three main components as shown in Figure~\ref{arch}: 1) Content-aware Feature Selector (CFS), 2) Position-aware Temporal Convolver (PTC), and 3) Content-dependent Multi-scale Aggregator (CMA). In this section, we will detail each component and how they overcome the drawback of CNNs.



\paragraph{\textbf{Content-aware Feature Selector}} First of all, the aim of Content-aware Feature Selector (CFS) is to adaptively select top-$k$ similar neighbouring frame-wise features. Given a series of extracted features from Frame Feature Extractor, CFS will select similar neighbours via three step as shown in Figure~\ref{feature_selector}: 

\textbf{(a)} With the extracted features $s^{T\times C'}$ from the backbone CNN model, we apply outer tensor product and softmax function to get a similarity matrix $d^{T\times T}$:

\begin{equation}
\begin{aligned}
d^{T\times T} = \textup{softmax}(\frac{s^{T\times C'}s^{{C'\times T}}}{C'})
\end{aligned}
\label{eqn:equation5}
\end{equation}

\noindent where the diagonal elements in $d^{T\times T}$ represent similarities of the features themselves.

\textbf{(b)} To ensure neighbors are going to be selected instead of the far-away ones, we only consider a range $[t-k, t+k]$ for a specific feature $s_t\in{s^{T\times C'}}$ to keep local semantic consistency. Outside that range, we replace the similarity scores with $-inf$. Mathematically, the selecting criterion for $s_t$ becomes as:

\begin{equation}
\begin{aligned}
d_{t,j} = \begin{cases} 
d_{t,j}, & j \in [\max(0, t-k), \min(t+k, T)]\\
-inf, & \text{others} 
\end{cases}
\end{aligned}
\label{eqn:equation6}
\end{equation}

\textbf{(c)} Notably, we denote the selected $k$ neighbours as $\mathbb{K}_t$ (i.e., \textbf{L}ocally \textbf{S}imilar \textbf{R}egion) for a specific feature $s_t$. Finally, we obtain a sub-series of enhanced frame-wise representations via temporal similarity as:

\begin{equation}
\begin{aligned}
s^{T\times k \times C'} = \mathcal{F}_{topk}(s^{T\times C'}, \mathbb{K})
\end{aligned}
\label{eqn:equation7}
\end{equation}

\noindent where the $k$ neighbors are sorted based upon the time steps in ascending order.

Since different glosses usually correspond to different frame lengths, we propose multi-scale locally similar regions to accommodate such phenomenon rather than a fixed one. Here, we use three region sizes $k_1$, $k_2$ and $k_3$, and accordingly obtain three set of selected frames, $s^{T\times k_1 \times C'}$, $s^{T\times k_2 \times C'}$ and $s^{T\times k_3 \times C'}$.

\paragraph{\textbf{Position-aware Temporal Convolover}} To further facilitate the temporal consistency between the selected frames, we then use temporal convolutional layers to sequentially fuse the selected features. However, according to Islam \textit{et. al}~\cite{Islam2020HowMP}, CNN layers are insensitive to what they are convolving in an image. And thus 1D-CNNs suffers from the same problem that they are insufficient to fully learn temporal dependency in between neighbouring frames. To encode the relationship of temporal position between frames, we adopt the relative position encoding~\cite{Shaw2018SelfAttentionWR,Dai2019TransformerXLAL,Yan2019TENERAT} to inject temporal dependency into the sign feature learning.

Concretely, we first calculate the relative positions w.r.t the center feature and map them into trainable embedding vectors. Then the relative position embeddings are added to the selected features, resulting in position-aware sign features:



\begin{equation}
\begin{aligned}
{\hat {s}^{T\times k \times C'}}_{t,p} = {s^{T\times k \times C'}_{t,p}} + \phi(p-t)
\end{aligned}
\label{eqn:equation8}
\end{equation}

\noindent where ${s^{T\times k \times C'}_{t,p}}$ denotes the $p^{th}$ selected feature in the region around the center $s_{t}$, and $p-t$ is the relative position value between the $p^{th}$ feature and the center $s_t$.

With a specific scale of selected features, we apply one-dimensional convolution layers to aggregate the position-aware sign features. The enhanced sign feature $\tilde{s}\in R^{T\times C'}$ is represented as:

\begin{equation}
\begin{aligned}
{\tilde{s}^{T\times C'}} = MaxPool(Relu(LN(Conv1d(\hat{s}^{T\times k \times C'}))))
\end{aligned}
\label{eqn:equation9}
\end{equation}

\noindent where Conv1d stands for 1D-CNN and LN denotes Layer Normalization~\cite{Ba2016LayerN}. In order to learn more expressive features, we add two linear layers $W_1\in \mathbb{R}^{C'\times C'}$, $W_2 \in \mathbb{R}^{C'\times C'}$ with a GELU activation~\cite{Hendrycks2016GaussianEL} in between as follows:

\begin{equation}
\begin{aligned}
{s'}^{T\times C'} = W_2 \cdot \text{GELU}(W_1 \tilde{s}^{T\times C'} +b_1) + b_2
\end{aligned}
\label{eqn:equation10}
\end{equation}

\noindent where $s'\in \mathbb{R}^{T\times C'}$ denotes enhanced sign features with local similar features. 

Accordingly, the same operations~\ref{eqn:equation8}, \ref{eqn:equation9}, \ref{eqn:equation10} are applied to all the scales ($k_1$, $k_2$ and $k_3$) of selection regions. Finally, we concatenate the three resulting enhanced features into  ${s'}^{T\times 3 \times C'}$, which still needs to be fused into a 2D matrix for the following recognition.

\paragraph{\textbf{Content-dependent Multi-scale Aggregator}} In order to fuse the three-scaled features  ${s'}^{T\times 3 \times C'}$, we are inspired by Chen~\textit{et. al}~\cite{Chen2020DynamicCA} and re-weight the three scales content-dependently at each time step.  

To be specific, we introduce a linear projection and map them into a weight distribution $\alpha$, which is also shown in the right side of Figure~\ref{mtsf}:


\begin{equation}
\begin{aligned}
\alpha^{T\times 3} = \text{softmax}(W s^{T\times C'} + b)
\end{aligned}
\label{eqn:equation11}
\end{equation}

\noindent where  $W \in \mathbb{R^{C'\times 3}}$ and $b\in \mathbb{R^{3}}$. And then, we dynamically aggregate these three scales using the weight distribution $\alpha$ as:


\begin{equation}
\begin{aligned}
s_o^{T\times C'} = a^{T\times 3} \cdot s_{o}^{T\times 3\times C'} 
\end{aligned}
\label{eqn:equation12}
\end{equation}

\noindent where the resulting feature $s_o$ is the final aggregated representation. With the learned weight distribution, our model is able to dynamically emphasize on different scale at different time step, e.g., if a gloss is across a wide range, and the model will pay pay more attention onto the bigger scale size of selection region, and vice versa.

\paragraph{\textbf{Summary of mLTSF}} On the one hand, our mLTSF-Net (CFS + PTC + CMA) is induced to capture temporal local similarity in sign language, on the other hand it is also able to enhance temporal consistency of every fused representation.



\subsection{Sequence Learning}
It is hard to train a cSLR model, since the source and target sequences usually have different lengths. Besides, there is no specific alignment between the video frames and the corresponding signing glosses.

To tackle the problem, we use Connectionist Temporal Classification (CTC)~\cite{Graves2006ConnectionistTC} to optimize our model for cSLR. Specifically, CTC introduces a designed cost function for sequence prediction, where the sequence is purely unsegmented or is weakly segmented. Besides, CTC extends the sign vocabulary with a special token ``blank", and we name this vocabulary as $V$. Further, the token ``blank" is expected to label the silence or transition frame that may exist in between output sign labels.

To transform the gloss features into gloss labels, we adopt a fully-connected layer \(\mathcal{F}_{fc}\) on the outputs from the 2nd level of gloss feature encoder in Equation~\ref{eqn:equation4}. Mathematically, we obtain the normalized probabilities of predicted  labels as:

\begin{equation}
\begin{aligned}
p^{T'\times V} = \mathcal{F}_{fc}(g_{2}^{T'\times C^{o}})
\end{aligned}
\label{eqn:equation13}
\end{equation}


Moreover, CTC defines a mapping function \(\mathcal{B}\): \(\mathcal{Y}^L = \mathcal{B}(\pi^{T'})\) by removing ``blank" tokens and collapsing repetitions. Note that  \(L \le T'\) and \(\pi\) is a possible alignment sequence.  During training, the objective of CTC is to minimize the negative log-likelihoods of all possible alignments, \(\mathcal{B}^{-1}(\mathcal{Y}) = \{\pi: \mathcal{B}(\pi)=\mathcal{Y}\}\):

\begin{equation}
\begin{aligned}
\mathcal{L}_{ctc}(\mathcal{X, Y}) = -\sum_{\small{\pi \in \mathcal{B}^{-1}(\mathcal{Y})}}\log{p(\mathcal{Y}|\mathcal{X})}
\end{aligned}
\label{eqn:equation14}
\end{equation}

Overall, we train the whole model by minimizing the loss as follows:

\begin{equation}
\begin{aligned}
\mathcal{L} = \mathcal{L}_{ctc} + \lambda \lVert W \rVert_{2}
\end{aligned}
\label{eqn:equation15}
\end{equation}

\noindent where \(\lambda\) is the \(L_2\) regularization coefficient.

\section{Model Configuration}\label{model design}

\begin{table}[t]
\renewcommand\arraystretch{1.2}
\centering
\resizebox{0.7\textwidth}{!}{
\begin{tabular}{|c:c|}
\hline
\multicolumn{2}{|c|}{\textbf{Frame Feature Extractor}} \\
\multicolumn{2}{|c|}{CNN backbone} \\
\hline \hline
\multicolumn{2}{|c|}{\textbf{mLTSF Module}} \\
\multicolumn{2}{|c|}{Content-aware Feature Selector (\(k\)=\{16,12,8\})} \\
\hdashline
\multirow{5}{*}{Position-aware Temporal Convolver} &  RPE (C512) \\
& Conv1D (F3-S1-P1-C512) \\
& Conv1D (F3-S1-P1-C512) \\
& AdaptiveMaxPool (M1)\\
& Linear + GLEU + Linear \\
\hdashline
\multicolumn{2}{|c|}{Content-dependent Multi-scale Aggregator} \\
\hline \hline
\multicolumn{2}{|c|}{\textbf{1st Level of Gloss Feature Encoder}} \\
\multicolumn{2}{|c|}{Conv1D (F5-S1-P2-C512)} \\
\multicolumn{2}{|c|}{Maxpool1D (M2-S2)} \\
\multicolumn{2}{|c|}{Conv1D (F5-S1-P2-C512)} \\
\multicolumn{2}{|c|}{Maxpool1D (M2-S2)} \\
\hline \hline
\multicolumn{2}{|c|}{\textbf{2nd Level of Gloss Feature Encoder}} \\
\multicolumn{2}{|c|}{Conv1D (F3-S1-P1-C1024)} \\
\hline
\end{tabular}}
\caption{The configuration of our proposed architecture.}
\label{model setting}
\end{table}

\begin{table}[t]
\renewcommand\arraystretch{1.1}
\centering
\begin{tabular}{ll}
\hline
\multicolumn{2}{c}{\textbf{CNN Backbone}} \\
\hline
Conv2D, \(3\times 3,1,1,32\) & MaxPool2D, \(2\times 2, 2\) \\
Conv2D, \(3\times 3,1,1,32\) & \\
Conv2D, \(3\times 3,1,1,64\) & MaxPool2D, \(2\times 2, 2\) \\
Conv2D, \(3\times 3,1,1,64\) & \\
Conv2D, \(3\times 3,1,1,128\) & MaxPool2D, \(2\times 2, 2\) \\
Conv2D, \(3\times 3,1,1,128\) & \\
Conv2D, \(3\times 3,1,1,256\) & MaxPool2D, \(2\times 2, 2\) \\
Conv2D, \(3\times 3,1,1,256\) & \\
Conv2D, \(3\times 3,1,1,512\) & MaxPool2D, \(2\times 2, 2\) \\
Conv2D, \(3\times 3,1,1,512\) & \\
Global Average Pooling \\
\hline
\end{tabular}
\caption{The settings of our backbone model, which is used to extract spatial features from the input sign language frames. The parameters of 2D convolution layer are denoted as "Conv2D, filter size \(\times\) filter size, stride, padding, output channels".}
\label{backbone}
\end{table}


In this section, we will provide more details of our model in a left-to-right order as illustrated in Figure~\ref{arch}. The overall configurations of our proposed architecture are presented in Table~\ref{model setting}. 

In terms of the feature extractor shown as in Figure~\ref{arch}, we follow the setting of Cheng \textit{et al.} \cite{Cheng2020FullyCN}, where the parameters are listed in Table~\ref{backbone}. Besides, we also add batch normalization~\cite{ioffe2015batch} to accelerate training after every convolution layer.

Followed right after by the frame feature extractor, our proposed mLTSF module is consisted of three components, 1) Content-aware Feature Selector (CFS), 2) Position-aware Temporal Convolver (PTC), and 3) Content-dependent Multi-scale Aggregator (CMA). The CFS component is used to capture similarity in multi-scale regions, here we set the sizes of three selection regions for CSF as \(k_1=16\), \(k_2=12\), and \(k_3=8\). We use PTC component to sequentially convolve a set of selected similar features. Here, we stack two 1D-CNNs for PTC, where the filter size is set as as 3, the stride size as 1, and the padding size as 1. Also, we add Layer Normalization~\cite{Ba2016LayerN} after the 1D-CNN layer. Afterwards, the CMA component just follows equation~\ref{eqn:equation11} and \ref{eqn:equation11} to fuse the three scales. 


The 1st level of gloss feature encoder is stacked with two same compounds (where each compound is consisted of a 1D-CNN and a TP (temporal max pooling)). Specifically, we set the filter size of 1D-CNNs as 5, stride size as 1, the padding size is 2, and the channel number is 512. For the sake of consistency, we name such setting as F5-S1-P2-C512. Besides, the TP layer has a pooling size of 2 and the stride size of 2. Similarly, we define this setting as M2-S2. Therefore, the overall receptive field of the 1st level of gloss feature encoder is equal to 16, which is ensured to be equivalent to $k_1=16$, the largest size of our selection region. Therefore, the original sequence length $T$ of video frames is reduced to $T'=\frac{T}{4}$. Accordingly, the length of a gloss unit is scaled down by four times. Then, we use a 1D-CNNs with the setting as F3-S1-P1-C1024 layer for the 2nd level of gloss feature encoder.


\section{Experiment}\label{Experiment}

\subsection{Dataset and Metrics}
We conduct experiment on RWTH-PHOENIX-Weather-2014 (RWTH)~\cite{Forster2014ExtensionsOT}, the benchmarking dataset of the continuous Sign Language Recognition. It is curated from a public television broadcast in Germany, where all signers wear dark clothes and perform sign language in front of a clean background. Specifically, this dataset contains \(\small{6,841}\) sign language sentences (around \(\small{80,000}\) glosses with a vocabulary of size \(\small{1,232}\)). Officially, the dataset is splitted with \(\small{5,672}\) training samples, \(\small{540}\) validation samples, and \(\small{629}\) testing samples. All videos have been pre-processed to a resolution of \(210 \times 260\) and \(\small{25}\) frames per second (FPS).

We evaluate the performance of our model using word error rate (WER), which is broadly adopted metric for cSLR:

\begin{equation}
\begin{aligned}
\text{WER} = \dfrac{\#\text{substitution} + \#\text{deletion} + \#\text{insertion}}{\#\text{words in reference}}
\end{aligned}
\label{eqn:equation16}
\end{equation}

\subsection{Implementation Details} 

We resize all the input frames to \(256\times 256\) for both training and testing. To augment the training dataset, we randomly crop the video frames to \(224\times 224\). In addition, we also scale up the training video sequence by +\(20{\%}\) and then scale down by -\(20{\%}\). During testing, we however crop the frames from the center, instead of randomness.


We use the Adam optimizer \cite{Kingma2015AdamAM} to optimize our model, where the initial learning rate is set to $10^{-4}$. After 40 epochs, we decay the learning rate by 0.5 every 5 epochs, where the decay rate $\lambda$ is $10^{-4}$. Overall, we train our model for 60 epochs in total on one NVIDIA Titan RTX GPU.

\begin{table}[!]
\centering
\smallskip
\resizebox{0.9\textwidth}{!}{
\begin{tabular}{lccccc}
\hline
\multirow{2}{*}{\textbf{Methods}} &  \multirow{2}{*}{\textbf{Iterative}} &
\multicolumn{2}{c}{\textbf{Dev (\%)}} &  \multicolumn{2}{c}{\textbf{Test (\%)}} \\
 & & del/ins & WER & del/ins & WER \\
 \hline
DeepHand~\cite{Koller2016DeepHH} & yes & 16.3/4.6 & 47.1 & 15.2/4.6 & 45.1 \\
Staged-Opt~\cite{Cui2017RecurrentCN} & yes & 13.7/7.3 & 39.4 & 12.2/7.5 & 38.7 \\
Re-Sign~\cite{Koller2017ReSignRE} & yes & - & 27.1 & - & 26.8 \\
CNN-LSTM-HMM~\cite{Koller2020WeaklySL} & yes & - & 26.0 & - & 26.0 \\
SBD-RL~\cite{Wei2021SemanticBD} & yes & 9.9/5.6 & 28.6 & 8.9/5.1 & 28.6 \\
DNF~\cite{Cui2019ADN} & yes & 7.8/3.5 & 23.8 & 7.8/3.4 & 24.4 \\
\hline
SubUNets~\cite{Camgz2017SubUNetsEH} & no & 14.6/4.0 & 40.8 & 14.3/4.0 & 40.7 \\
DeepSign~\cite{Koller2016DeepSH} & no & 12.6/5.1 & 38.3 & 11.1/5.7 & 38.8 \\
SFNet~\cite{Yang2019SFNetSF} & no & - & 38.0 & - & 38.1 \\
FCN~\cite{Cheng2020FullyCN} & no & - & 26.1 & - & 26.0 \\
FCN + GFE~\cite{Cheng2020FullyCN} & no & - & 23.7 & - & 23.9 \\
\hline
\textbf{mLTSF-Net} & no & 8.5/3.2 & 23.8 & 8.3/3.1 & \textbf{23.5} \\
\textbf{mLTSF-Net + GFE} & no & 7.9/2.7 & \textbf{22.9} & 7.6/3.0 & \textbf{23.0} \\
\hline
\end{tabular}}
\caption{The evaluation results on baseline models, where del and ins stand for deletion and insertion, respectively.}
\label{main results}
\end{table}

\subsection{Comparison Against Baselines}
To demonstrate the effectiveness of our work, we compare with several state-of-the-art models. The experimental results are summarized in Table~\ref{main results}.

We firstly examine the performance of iterative optimization methods, e.g.,  Stage-Opt~\cite{Cui2017RecurrentCN}, Re-sign~\cite{Koller2017ReSignRE} and DNF~\cite{Cui2019ADN}. The results are listed in the first block of Table~\ref{main results}. They resort to iteration in order to address the problem of weak supervision. They first train an end-to-end recognition model for alignment proposals, and then use the alignment proposals to finetune the frame feature extractor. However, these iterative methods are time consuming. Even worse, the proposed alignments for supervision are noisy, and thus tend to degrade the performance unavoidably. 


Among the non-iteration method as shown in the second block of Table~\ref{main results}, FCN~\cite{Cheng2020FullyCN} achieve a remarkable WER compared with the others. It is built upon an end-to-end fully convolutional neural network for cSLR. Further,  Cheng \textit{et.al.} \cite{Cheng2020FullyCN} introduce a Gloss Feature Enhancement (GFE) to enhance the frame-wise representation, where GFE is trained to provide a set of alignment proposals for the frame feature extractor. Notably, FCN+GFE exhibits great potentials for cSLR, and reduce the WER to a new low.

However, as discussed in previous sections, CNNs are insufficient to capture local similarities and insensitive to temporal order of a series of video frames. Built on FCN, our mLTSF-Net additionally introduces a mLTSF module to address the issues of CNNs for cSLR. As shown in the last two rows in Table~\ref{main results}, our model outperforms all the other baseline models on WER.  Combined with Gloss Feature Enhancement (GFE), our model can even achieve a better performance. Such results demonstrate the effectiveness of our proposed mLTSF module, and our model equipped with GFE achieves the state-of-the-art performance on the dataset of RWTH-PHOENIX-Weather-2014.




\section{Ablation Study and Analysis}

\begin{table}[!]
\centering
\smallskip
\resizebox{0.45\textwidth}{!}{
\begin{tabular}{lcc}
    \hline
    \multirow{2}{*}{mLTSF} & \multicolumn{2}{c}{WER} \\ 
    & DEV & TEST \\
    \hline
    None & 26.1 & 26.0 \\
    \hline
    \(k=8\) & 25.0 & 24.9 \\
    \(k=12\)  & 24.7 & 24.5  \\
    \(k=16\)  & 24.4 & 24.2 \\
    \(k=20\)  & 25.3 & 25.2 \\
    \hline
    \(k=\{8, 12\}\) & 24.3 & 24.5  \\
    \(k=\{8, 16\}\) & 23.9 & 24.2  \\
    \(k=\{12, 16\}\) & 24.0 & 23.9 \\
    \hline
    \(k=\{8, 12, 16\}\) & \textbf{23.8} & \textbf{23.5} \\
    \hline
    \end{tabular}
}
\caption{The effect of selection region sizes for mLTSF module.}
\label{size of mLTSF}
\end{table}

Compared with the FCN framework, our model has an additional module, i.e., mLTSF module, which is induced to adaptively fuse the local-temporal features, in order to capture local semantic consistency. In this section, we will compare closely with FCN to validate the effectiveness of our proposal.

\subsection{Single vs Multiple Scales}~\label{mtsf_vs_stsf}
The main idea of our proposed multi-scale selection regions is to dynamically adapt to different lengths of gloss. To demonstrate our idea, we evaluate the performance of our model with different combinations of selection region sizes in Equation~\ref{eqn:equation7} to answer the following two questions: 1) Are multiple scales are effectively necessary? 2) If necessary, how do the sizes of these scales influence the overall performance? 

The results are shown in Table~\ref{size of mLTSF}. The 2nd row (None), which represents the pure FCN and performs the worst, which does not contain our proposed mLTSF module. Such result indicates that our proposed mLTSF module is effective in improving the performance over the FCN framework. As we gradually increase the size from $k=8$ to $k=16$ as shown in Table~\ref{size of mLTSF},  the performance is promoted along with such size increase. Particularly, the performance reaches the best when we set $k$ as 16. This is not surprising since the average length of a gloss roughly equals to 16. However, when we set $k$ as 20, the performance becomes much worse. Such performance decline is mostly due to that 20 frames are usually contain more than on sign gloss, and similar frames which locate distantly might share little semantic consistency. As a result, the final representations tend to be indistinguishable for cSLR.




Comparing last four lines with a single scale as illustrated in Table~\ref{size of mLTSF}, we see that our model with multiple scales greatly outperforms than single scale. Especially when we use three scales with the setting as $k={8, 12, 16}$, our model achieves the lowest WER. Enhanced with multiple selection regions, our model is able to seize most salient scale to represent current time-step feature for downstream sign language recognition. On the other side, since different gloss is consists of different lengths, our model equipped with multiple scales are capable of accommodate such phenomenon. To sum up, this comparison demonstrates that it is recommendable to use multiple scales for mLTSF module with a setting as $k=\{8, 12, 16\}$.


\begin{figure}[t] 
  \begin{minipage}[b]{0.48\textwidth}
   \renewcommand\arraystretch{1.15}
    \centering 
    \begin{tabular}{ccc}
    \hline
    \multirow{2}{*}{Feature Selector} & \multicolumn{2}{c}{WER} \\ 
    & DEV & TEST \\
    \hline
    FCN &  26.1 & 26.0 \\
    \hline
    global-FS & 26.1 & 26.5 \\
    center-FS & 25.3 & 25.0 \\
    \hline
    CFS  & \textbf{24.4} & \textbf{24.2} \\
    \hline
    \end{tabular}
    \captionof{table}{The effect of feature selector.}
    \label{effect_of_cfs}
  \end{minipage}
  \begin{minipage}[b]{0.48\textwidth} 
    \centering
    \begin{tabular}{ccc}
    \hline
    \multirow{2}{*}{Method} & \multicolumn{2}{c}{WER} \\ 
    & DEV & TEST \\
    \hline
    w/o RPE  & 24.8 & 24.7  \\
    w/o TCNs & 25.7 & 25.6 \\
    mean pooling & 24.6 & 24.4  \\
    \hline
    PTC & \textbf{24.4} & \textbf{24.2} \\
    \hline
    \end{tabular}
    \captionof{table}{The effect of position-aware temporal convolver.}
    \label{effect_of_ptcn}
  \end{minipage} 
\end{figure}


\subsection{Analysis on Content-aware Feature Selector}\label{ana_cfs}
In our proposed mLTSF module, we introduce a content-aware feature selector (CFS) to dynamically select most similar features in a local region. To demonstrate its effectiveness, we conduct comparison study to explore two potential concerns: 1) Compared against a fixed local region, is it necessary to use content-aware mechanism to dynamically select neighbours based on the similarity? 2) Is it competent to select similar neighbours from all the whole sequence globally? To uncover these two doubts, we compare our proposed CFS with some feature selector variants as shown in Table~\ref{effect_of_cfs}. Note that FCN is the baseline model where there is no feature selector. For notation, "center-FS" stands for that the model simply picks its surrounding left and right neighbours, where there are 16 neighbours in total. Besides, "global-FS" stands for that the model selects top-$k$ similar features from the whole sequence globally, where $k$ is set as 16. Note that we use a single-scale receptive region to select top-$k$ similar features locally for a fair comparison with the variants, where $k$ is also set as 16.





The comparison results are listed in Table~\ref{effect_of_cfs}. It is surprising that global-CFS even degrades the performance compared with FCN. In such way, the final aggregated features become indistinguishable, since two similar frames but locates far away might not share much semantic consistency. On the contrary, center-FS brings a performance increase over the baseline FCN. This increase shows that the aggregation of neighbouring local region is a simple yet effective way,  where the model is able to further seize local semantic consistency. Moreover, our proposed CFS performs the best, which further demonstrates the importance of concentration of local similar features. This content-aware mechanism provides a channel to make full use of local semantic consistency. Besides, this comparison validates the effectiveness of our CFS component.





\subsection{Analysis on Position-aware Temporal Convolver}
To generate neighborhood-enhanced representation, we propose a position-aware temporal convolver to ensure the sequential consistency among each scale of selected features. Specifically, the position-aware temporal convolver (PTC) is  composed of a relative position encoding (RPE), temporal convolutions (TCNs), and an adaptive max-pooling layer. To testify the influence of each component, we apply ablation experiments to compare PTC with its variants. As shown in Table~\ref{effect_of_ptcn}, the first two variants are without RPE and TCNs, respectively. The third variant replaces the adaptive max-pooling layer with mean pooling. From such comparison, we can see that (1) without RPE, the model performance is reduced by 0.4/0.5 WER on DEV/TEST. This result validates that that relative position information is essential for sequence modeling, and confirms the assertion made by \cite{Islam2020HowMP} that the CNN filters tend to be insensitive with temporal dependency. (2) Without TCNs, we aggregate the selected features only using the max-pooling layer. The performance of our model drops significantly, showing that temporal convolution is still fundamental for local-temporal modeling. (3) We replace the adaptive max-pooling with mean-pooling, where the model tends to become insensitive to the salience. As a consequence, the performance is decreased by 0.2/0.2 WER. In summary, each component of PTC is indispensable, and our proposed PTC as a whole can effectively improve the performance.


\subsection{Analysis on Content-dependent Multi-scale Aggregator}
In order to seize multiple temporal granularities, we introduce a content-dependent multi-scale aggregator to fuse different scales of selection regions. To testify if it is necessary to fuse the three scales in a dynamic way, we conduct an experiment to compare our dynamic method with a static average one. The experimental results are shown in Table~\ref{effect_of_msagg}. The dynamic method significantly outperforms the average one in terms of WER. Since different gloss corresponds to different length, a static average method is insufficient to catch gloss-level salience. On the contrary, our dynamic method is able to fuse different scales with salience estimation, and generates a sequence of distinguishable features for downstream sign language recognition. Finally, this comparison further demonstrates that a content-dependent assembling method is an excellent approach for multi-scaled features fusion.

\begin{figure}[!] 
  \begin{minipage}[b]{0.48\textwidth}
    \centering
    \renewcommand\arraystretch{1.25}
    \begin{tabular}{c|cc}
    \hline
    \multirow{2}{*}{Aggregator Method} & \multicolumn{2}{c}{WER} \\ 
    & DEV & TEST \\
    \hline
    average & 24.3 & 23.9 \\
    dynamic & \textbf{23.8} & \textbf{23.5} \\
    \hline
    \end{tabular}
    \captionof{table}{The effect of content-dependent multi-scale aggregator.}
    \label{effect_of_msagg}
  \end{minipage} 
  \begin{minipage}[b]{0.48\textwidth} 
    \centering
    \begin{tabular}{c|cc}
    \hline
    \multirow{2}{*}{Method} & \multicolumn{2}{c}{WER} \\ 
    & DEV & TEST \\
    \hline
    non-local sparse attention & 43.0 & 46.6 \\
    local sparse attention & 38.2 & 36.6 \\
    single-LTSF & \textbf{24.4} & \textbf{24.2} \\
    \hline
    \end{tabular}
    \captionof{table}{Comparison between single-LTSF and sparse attention.}
    \label{tsf_vs_attention}
  \end{minipage} 
\end{figure}


\subsection{Our model vs Sparse Attention}
Our proposed content-aware feature selector is induced to select top-$k$ similar neighbours. This selection process is similar to Sparse Attention~\cite{Child2019GeneratingLS,Beltagy2020LongformerTL,Zhou2020InformerBE}, which have been shown great potentials in various tasks. They design various sparsity patterns, and find similar ones globally from the whole sequence. However, what is difference from our model is that they aggregate the selected features by weighted averaging. To verify whether our model is over designed for cSLR, we compare our model using only a single scale (top-16 similar features) with the sparse attention.

The experimental results are exhibited in Table~\ref{tsf_vs_attention}. When comparing the non-local with local sparse attention, the local one achieves a much better performance, where the local sparse attention is constrained to attend only in a local region. This performance promotion further confirms that similar frames which locate far away do not share much gloss-level consistency. Our model only using a single scale is named as single-LTSF, which is listed in the last row of Table~\ref{tsf_vs_attention}. Interestingly, single-LTSF outperforms significantly the other two comparable models. This improvement might due to many reasons, and we speculate that sparse attention which just uses weighted average is hard to ensure the temporal consistency with the selected sparse features, resulting an indistinguishable fusion for downstream sign language recognition. To sum up, the sparse attention could not model well the task of cSLR. On the contrary, it is more recommendable to use our proposed mLTSF-Net for the task of cSLR.





\begin{figure}
\centering
\includegraphics[width=\textwidth]{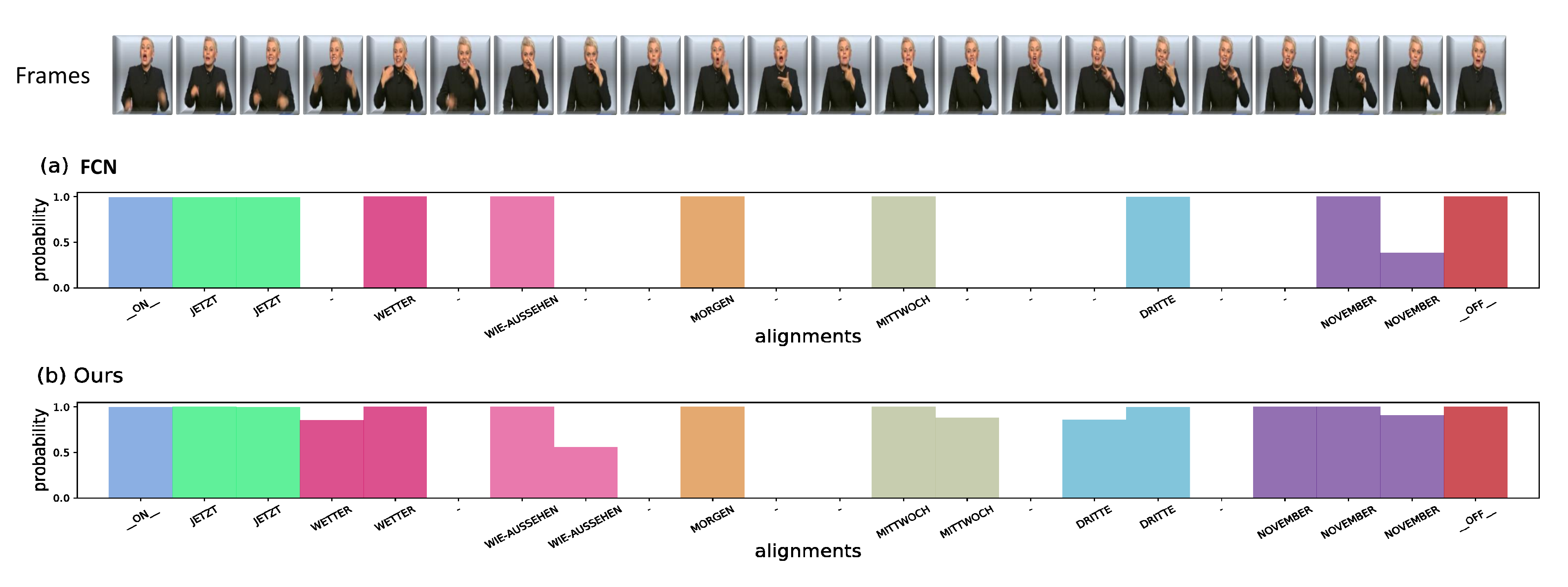}
\caption{Alignments for a training example by FCN and Ours. The x-axis represents the alignment with the highest probability among all possible underlying alignments, where "-" represents "blank" symbol. While the y-axis represents the corresponding probabilities.}
\label{confidence} 
\end{figure}

\subsection{Case study}
To better understand the effect of our mLTSF-Net, we visualize the alignments of FCN and our model (FCN + mLTSF) in Figure~\ref{confidence}. Particularly, we see that FCN is more likely to align  ``blank" tokens to the frames at the boundaries. For example, the frames around ``NOVEMBER" (purple bars) are very similar. However, FCN does not capture such similarity, and assign a ``blank" or gives a very low probability for ``NOVEMBER". In contrast, our mLTSF-Net performs much better. Thanks to mLSTF module, our model is able to capture such similarity, and gives a correct alignment or even assign a much higher probability. This comparison further demonstrates that our mLTSF module is able to capture local semantic consistency. This is the most important factor for performance improvement.

\section{Conclusion}\label{conclusion}
In this paper, we identify the drawback of fully convolutional network for cSLR that CNNs would tend to become agnostic to similarity and temporal consistency, and thus could not model well local semantic consistency in sign language. To tackle it, we propose a mLTSF-Net which is built upon the FCN network to fuse temporal similar frames with multiple granularities, in order to keep consistent semantics in between neighboring frames. On the standard benchmarking RWTH dataset, we demonstrate that our approach outperforms several baseline models. The extensive ablation study proves the effectiveness of our proposed mLTSF-Net for cSLR.

 \bibliographystyle{elsarticle-num} 
 \bibliography{cas-refs}





\end{document}